# Comparison of Path Planning Algorithms for Autonomous Vehicle Navigation Using Satellite and Airborne LiDAR Data

Chang Liu [1,2,*], Zhexiong Xue [1], Tamas Sziranyi [1,2]

[1] Department of Networked Systems and Services, Faculty of Electrical Engineering and Informatics, Budapest University of Technology and Economics, Műegyetem rkp. 3, H1111 Budapest, Hungary

[2] Machine Perception Research Laboratory, HUN-REN Institute for Computer Science and Control (SZTAKI), H-1111 Budapest, Kende u. 13-17, Hungary

changliu@hit.bme.hu; liu.chang@sztaki.hun-ren.hu

*Abstract*—Autonomous vehicle navigation in unstructured environments, such as forests and mountainous regions, presents significant challenges due to irregular terrain and complex road conditions. This work provides a comparative evaluation of mainstream and well-established path planning algorithms applied to weighted pixel-level road networks derived from high-resolution satellite imagery and airborne LiDAR data. For 2D road-map navigation, where the weights reflect road conditions and terrain difficulty, A*, Dijkstra, RRT*, and a Novel Improved Ant Colony Optimization Algorithm (NIACO) are tested on the DeepGlobe satellite dataset. For 3D road-map path planning, 3D A*, 3D Dijkstra, RRT-Connect, and NIACO are evaluated using the Hamilton airborne LiDAR dataset, which provides detailed elevation information. All algorithms are assessed under identical start and end point conditions, focusing on path cost, computation time, and memory consumption. Results demonstrate that Dijkstra consistently offers the most stable and efficient performance in both 2D and 3D scenarios, particularly when operating on dense, pixel-level geospatial road-maps. These findings highlight the reliability of Dijkstra-based planning for static terrain navigation and establish a foundation for future research on dynamic path planning under complex environmental constraints.

## I. Introduction

Autonomous vehicle (AV) navigation has witnessed remarkable progress in recent years, particularly in structured urban environments where high-definition maps and lane-level localization are available. However, deploying AV systems in unstructured or semi-structured environments—such as forests, mountainous regions, or disaster zones—remains a significant challenge [1]. These terrains often feature irregular topography, undefined or partially visible roads, and a dynamic array of natural obstacles, all of which pose serious difficulties for reliable path planning. Most traditional path planning algorithms are typically developed and tested in synthetic or simulation-based environments [2]. These simulations, while useful for algorithm design, often lack the real-world complexity present in pixel-level geospatial data derived from remote sensing. In unstructured areas, where high-definition maps are not available, remote sensing becomes a crucial tool for environment perception and map generation [3].

High-resolution satellite imagery and airborne LiDAR (Light Detection and Ranging) data offer a powerful alternative by enabling the extraction of detailed road networks and terrain features from a bird's-eye view. Satellite imagery provides wide-area visual information, including surface textures, vegetation cover, and road morphology, making it suitable for identifying accessible routes, land use patterns, and potential obstructions. In contrast, LiDAR offers precise 3D elevation data, capturing subtle changes in terrain elevation and slope gradients, which is essential for generating terrain-aware maps that support obstacle avoidance and safe path planning in rugged or forested environments. Building on our previous work [4,17], we further demonstrated that satellite imagery can be used not only to extract road networks but also to evaluate road surface conditions at the pixel level. Through this fine-grained analysis, each pixel representing a road segment was assigned a weight based on surface material, condition level, segment length, throughput potential, and connectivity. This resulted in the generation of highly informative weighted road-maps, which reflect the true navigational cost and accessibility of each route segment in complex terrains.

Despite the growing availability of such data sources, systematic comparisons of widely used path planning algorithms based on satellite and LiDAR data remain limited. Most existing studies focus on structured, rule-based urban settings and rarely address the integration of 2D and 3D planning in unstructured or elevation-rich environments. Moreover, few works compare the performance of traditional and metaheuristic algorithms on realistic, pixel-level road-maps extracted from remote sensing data. This work aims to address the above limitations by proposing a unified framework for the comparative evaluation of popular path planning algorithms in the context of autonomous navigation over remote sensing-derived maps. The specific objectives of this work are:

- To implement and assess a set of well-established path planning algorithms, including A* [5], Dijkstra [6], RRT* [7], and a Novel Improved Ant Colony Optimization Algorithm (NIACO) [8], on 2D weighted road-map derived from satellite imagery;

- To evaluate the corresponding 3D counterparts—3D A*, 3D Dijkstra, RRT-Connect [9], and Novel Improved Ant Colony Optimization Algorithm (NIACO)—on terrain-aware road-maps constructed using airborne LiDAR data;
- To use consistent start and goal configurations across all tests to enable fair comparisons based on path cost, computational time, and memory usage;
- To investigate the stability and scalability of these algorithms when applied to dense, pixel-level geospatial data, with the aim of supporting future developments in dynamic and real-time path planning under complex environmental constraints.

## II. RELATED WORK

### A. Data Sources and Descriptions

To construct realistic and high-resolution road-maps for AV path planning, we employed both satellite imagery and airborne LiDAR point cloud data. For 2D path extraction, we adopted the publicly available DeepGlobe Road Extraction Dataset [10], which was released in 2018. This dataset, based on DigitalGlobe Vivid Images, includes satellite images from regions such as Thailand, Indonesia, and India. The original geotiff image size is 19584×19584 pixels with a ground resolution of 50 cm/pixel. After pre-processing, the images are cropped to 1024×1024 patches for training and testing, and consist of standard RGB channels. However, these images only offer 2D surface information, which is insufficient for our 3D path planning tasks. To incorporate terrain elevation, we utilized high-resolution point cloud data from OpenTopography, specifically the Hamilton, Waikato, New Zealand 2023 dataset [11]. This dataset provides Digital Elevation Models (DEM) at a resolution of 1 meter per pixel. It includes various meshing strategies such as minimum (Zmin), maximum (Zmax), and mean (Zmean) elevation values for each grid point, offering flexibility in terrain modeling. Each pixel in the generated DEM represents a 1m × 1m ground area in geographic coordinates. We used CloudCompare [12], an open-source point cloud processing tool, to visualize and extract Z-axis elevation values from the terrain data. These Z-values were transformed into grayscale maps representing terrain altitudes, which were then aligned with the 2D weighted roadmap to enable elevation-aware path planning. By combining pixel-level 2D road information from DeepGlobe with elevation data from LiDAR-derived DEMs, we built a foundation for integrated 2D and 3D path planning in complex environments.

### B. Road-map Extraction and Road Surface Conditions Evaluation

To accurately extract road structures and evaluate road usability in complex environments, we adopted *D-LinkNet* [13] for 2D road-map segmentation. *D-LinkNet*, a powerful semantic segmentation network designed for road extraction tasks, demonstrates superior performance in maintaining road connectivity even in occluded or fragmented satellite imagery. Its encoder-decoder architecture with dilated convolutions effectively captures both global and local context, which significantly improves the continuity and precision of the extracted road network. Building upon our previous work [4], which focused on road surface material classification and condition assessment, we integrated additional environmental and surface-level features into the pre-processing stage. Specifically, we utilized NDVI (Normalized Difference Vegetation Index) [14] to analyze the vegetation coverage and infer surface material properties, such as distinguishing between soil, gravel, and paved roads. The NDVI values, combined with texture and brightness features extracted from multi-spectral imagery, enabled us to classify road segments and assess their condition—ranging from well-maintained to degraded surfaces. Furthermore, by incorporating traffic-related features such as estimated throughput capacity and width consistency, we enriched the extracted road network with semantic weights. Each road segment extracted by *D-LinkNet* was assigned a corresponding weight that reflects its suitability for traversal, considering surface material, structural integrity, and environmental impact. The resulting weighted road-map provides a more realistic and functional representation of the terrain, supporting downstream tasks such as dynamic path planning and emergency route optimization.

## III. METHODOLOGY

### A. Framework of Proposed System

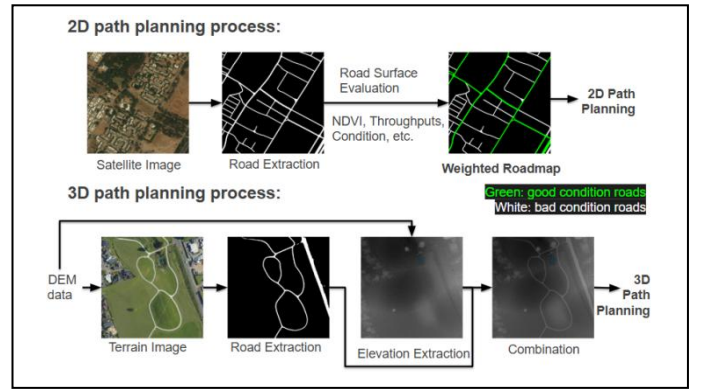

Figure 1.  Process of the 2D &3D Path Planing Framework

Figure 1 illustrates the overall framework of our autonomous vehicle (AV) navigation system, which includes two modules: 2D path planning using satellite imagery and 3D path planning incorporating elevation data. For the 2D module, we use RGB images from the DeepGlobe Road Extraction Dataset [10] to extract initial road-maps via *D-LinkNet*. To evaluate road surface conditions, NDVI is computed by adding an extra spectral channel, enabling the classification of roads (e.g., paved vs. unpaved) and surface quality estimation [4]. Each road segment is then assigned a weight reflecting its traversability, forming a weighted 2D road-map. Four path planning algorithms are applied from the same starting point to compare performance under different road conditions. The 3D module enhances the 2D network with elevation data from the 2023 Hamilton, New Zealand airborne LiDAR dataset [11], which provides 1m-resolution Digital Elevation Models. Elevation values are used to build a 3D terrain map, allowing slope-aware cost modeling. The same algorithms are run on this elevation-augmented map to assess their effectiveness in complex terrain.

*B. 2D Path Planning Algorithms*

In the path planning of 2D road-maps, we incorporated weight information and made appropriate modifications to the adopted algorithms. Dijkstra's algorithm [5], a single-source shortest path algorithm, is suitable for weighted graphs. Its basic principle involves continuously expanding the set of nodes with known shortest paths while updating the estimated shortest path values of neighboring nodes. In this work, since path weights vary based on road condition information, Dijkstra's algorithm is particularly applicable. Equation (1) demonstrates the core mechanism of the Dijkstra algorithm. For each neighboring node u:

$$d[v] = \min(d[v], d[u] + w(u,v)) \qquad (1)$$

where d[v] is the path cost from the source node to node v, and w(u,v) is the edge weight between nodes u and v. The algorithm proceeds in a greedy manner by prioritizing nodes with the smallest path cost until all nodes have been processed or the priority queue is empty.

The A* algorithm [6] is an extended version of the Dijkstra algorithm that incorporates a heuristic function to guide the search direction, thereby improving efficiency. Its cost function is defined as:

$$f(n) = g(n) + h(n) \qquad (2)$$

Here, *g(n)* represents the cost from the starting point to the current node n, and is typically calculated as:

$$g(n) = g(parent(n) + w(parent(n), n) \qquad (3)$$

where w(parent(n),n) is the weight of the edge between the parent node and the current node. The term *h(n)* is the heuristic estimated cost, often calculated as the Euclidean distance [15] from the current node to the goal node:

$$h(n) = \sqrt{(x_n - x_{goal})^2 - (y_n - y_{goal})^2} \qquad (4)$$

RRT* (Rapidly-Exploring Random Tree Star) [7] is an asymptotically optimal sampling-based path planning algorithm that rapidly explores the search space through random sampling and continuously refines the path to reduce cost. In each iteration, a random point is sampled from the environment:

$$x_{rand} \sim X \qquad (5)$$

Then it finds the nearest node $x_{nearest}$ in the current tree and extends a new node $x_{new}$ towards it, with the extension distance limited by a maximum step size Δ. In this work, the path cost uses edge weights $w(x_i, x_j)$ from the weighted road network instead of Euclidean distance. Specifically, the cost of reaching the new node is calculated as:

$$c(x_{new}) = c(x_{nearest}) + cost(x_{nearesti}, x_{new}) \qquad (6)$$

Where $cost(x_{nearesti}, x_{new}) = w(x_{nearesti}, x_{new})$ represents the weight between the two nodes, incorporating factors such as terrain difficulty or road quality. RRT* employs a rewiring mechanism that searches for lower-cost connections in the neighborhood around the new node:

$$x_{parent} = \underset{x \in \mathcal{X}_{near}}{\arg\min} (c(x) + w(x, x_{new})) \qquad (7)$$

If the new node $x_{new}$ can reduce the path cost of neighboring nodes, their parent nodes are updated to optimize the path. This process not only ensures rapid initial path discovery but also enables convergence towards the global optimum, making RRT* well-suited for complex, real-world environments modeled by weighted road networks.

The Novel Improved Ant Colony Optimization (NIACO) algorithm [8] is inspired by the foraging behavior of ants, which find shortest paths by laying down and following pheromone trails. NIACO is an enhanced version of the traditional ant colony optimization (ACO) [16] designed to address some key limitations such as slow convergence and tendency to get trapped in local optima. NIACO introduces a dynamic state transition strategy with a parameter $q_0$ that balances exploration and exploitation: early iterations favor a greedy approach to speed up convergence, while later iterations maintain randomness to preserve diversity in the search. Additionally, NIACO refines the pheromone update rules by incorporating dynamic evaporation rates and gradually decreasing pheromone increments, which help prevent premature convergence and encourage a more effective global search. In our case, the NIACO algorithm is integrated with a weighted road network where path costs influence the pheromone levels and heuristic functions. This weighting guides the ants to preferentially explore lower-cost routes, enabling the algorithm to find more practical and optimized paths tailored to real-world constraints.

*C. 3D Path Planning Algorithms*

Based on the introduction of the 2D path planning algorithms, this subsection focuses on the presentation of 3D path planning algorithms. Even in three-dimensional space, the fundamental theoretical framework of the A* algorithm remains applicable. By combining the cost function *g(n)* and the heuristic function *h(n)*, it is still possible to search for the shortest path in 3D environments. Nevertheless, the heuristic function *h(n)* needs to be designed by incorporating terrain elevation information. The 3D A* algorithm improves the path cost calculation by integrating elevation data. In addition, we also introduce Dijkstra's algorithm into the 3D context. Unlike the 2D scenario, the 3D implementation of Dijkstra's algorithm no longer relies solely on pixel points of a flat map but places greater emphasis on elevation information to achieve more accurate path planning.

Compared to 2D path planning where RRT* is commonly used, we adopt RRT-Connect [9] for 3D scenarios due to its superior efficiency and lower randomness. RRT-Connect grows two trees from the start and goal points respectively and connects them using a greedy heuristic, making it more suitable for high-dimensional single-query planning. Unlike RRT*, it does not depend on the underlying structure of the state space and avoids the long convergence time and non-optimality issues associated with probabilistic methods. Its original application in robotic manipulation has proven effective for 3D collision-free path planning [8]. To extend the NIACO algorithm to 3D space, we simply incorporate elevation data into the state transition probability. This allows the ants' decision-making to account for not only pheromone intensity and distance heuristics, as in 2D, but also movement cost related to terrain elevation. The added dimension makes the search more realistic for environments where altitude variation significantly affects traversal cost, enhancing the suitability of NIACO for AV path planning in complex terrains.

## IV. Experimental Results

### A. Comparative Evaluation of 2D Path Planning Algorithms on Weighted Road Maps

Figure 2 illustrates the results obtained by four 2D path planning algorithms. The path generated by the A* algorithm is shown in blue, Dijkstra's in gold, RRT* in red, and NIACO in light blue. The start and goal positions are listed below each subfigure. The left column of Figure 2 shows the complete path generated by each algorithm, while the right column highlights the critical branching points in the routes. From sub-figure (a), it can be observed that the A* and Dijkstra algorithms are nearly identical in their path selection, both opting for a green low-cost route. In contrast, the RRT* algorithm fails to make the correct choice, and the path generated by NIACO exhibits noticeable jitter, which negatively affects its overall performance. Overall, Dijkstra provides the smoothest path with minimal jitter. In sub-figure (b), all algorithms except RRT* make similar decisions at the fork, while RRT* chooses an incorrect direction, resulting in a significantly higher path cost. In sub-figure (c), Dijkstra and NIACO adopt similar strategies, whereas A* and RRT* diverge in their decisions. Among all, RRT* produces the least optimal route. Although RRT* is theoretically capable of finding the optimal path, its performance is hindered by the randomness inherent in its sampling strategy, leading to instability in path selection. Introducing heuristic mechanisms in future work may help mitigate this issue.

The corresponding performance metrics for Figure 2 are presented in Table 1. From the first set of paths, it is clear that the RRT* algorithm deviates significantly from the optimal path, resulting in much higher path costs compared to the other three algorithms. However, due to its low data maintenance requirements during execution, RRT* has the lowest memory consumption. The A* and Dijkstra algorithms produce similar results, though A* takes more time and requires more memory due to the extensive data it maintains during operation. In the second path planning scenario, all algorithms except RRT* generate roughly the same path. The tables also show that A* has the shortest time consumption, though the path cost is higher than that of Dijkstra. NIACO exhibits the longest runtime and sub-optimal path cost in this case. In the last planning result (sub-figure c of Figure 2), RRT* once again deviates from the optimal route. While it still benefits from low memory usage, the path quality is poor. Dijkstra achieves the lowest path cost, and although A* is the fastest, its higher path cost and memory usage remain concerns.

In summary, although RRT* theoretically offers optimal path finding capabilities and advantages in runtime and memory usage, its random sampling nature results in performance instability. NIACO requires many iterations to converge, and in this experimental setting, maintaining path quality comes at the cost of high time consumption, leading to poor overall performance. The A* algorithm can quickly find sub-optimal paths with reasonable time efficiency, but its path cost is higher than that of Dijkstra. Dijkstra consistently delivers stable performance, reliably finding optimal or near-optimal paths while maintaining acceptable runtime and memory usage.

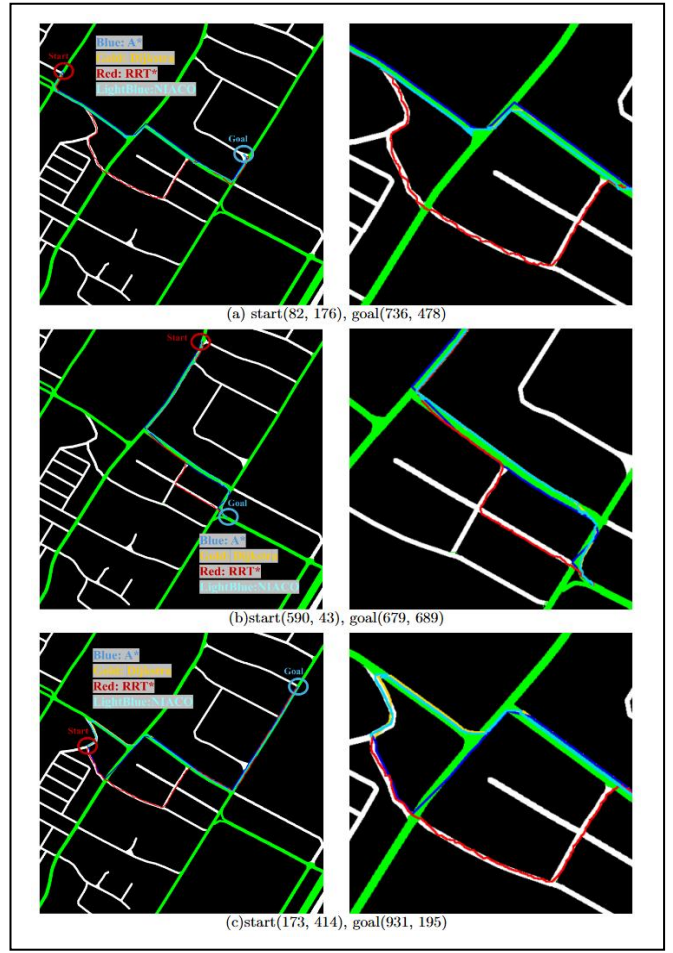

Figure 2. Comparison of 2D Path Planning Algorithms

TABLE I. Performance Comparison of 2D Path Planning Algorithms

| Metric | Algorithm | Path 1 | Path 2 | Path 3 |
|---|---|---|---|---|
| Path Cost | A* | 2383.8 | 931.0 | 15403.6 |
| | Dijkstra | 2470.8 | 911.8 | 12492.0 |
| | RRT* | 63269.8 | 23757.8 | 50967.4 |
| | NIACO | 10795.6 | 2420.2 | 33589.4 |
| Computation Time (s) | A* | 46.34 | 8.07 | 11.39 |
| | Dijkstra | 28.86 | 13.62 | 23.82 |
| | RRT* | 28.92 | 81.18 | 94.67 |
| | NIACO | 179.91 | 125.63 | 145.70 |
| Memory Usage (MB) | A* | 31.12 | 31.14 | 31.16 |
| | Dijkstra | 22.40 | 16.49 | 20.93 |
| | RRT* | 3.54 | 3.99 | 4.11 |
| | NIACO | 15.93 | 14.79 | 15.99 |

## B. Comparative Evaluation of 3D Path Planning Algorithms on Weighted Terrain Maps

In the 3D path planning evaluation results shown in Figure 3, the path generated by the 3D A* algorithm is displayed in blue, the 3D Dijkstra algorithm in gold, the 3D RRT-Connect algorithm in red, and the NIACO algorithm in light blue. In regions with higher terrain, the grayscale value is higher and the pixel color appears closer to white. As shown in sub-figure (a), the 3D RRT-Connect algorithm selects a path that passes through high terrain, resulting in a correspondingly higher path cost. The paths chosen by the 3D A* and 3D Dijkstra algorithms are generally similar; however, the path planned by the 3D A* algorithm lacks smoothness, leading to an unnecessary increase in path cost. This is because the 3D A* algorithm uses local sampling to calculate the average gradient of neighboring pixels for planning.

However, the grayscale image in the figure contains a certain amount of noise, which originates from the point cloud data and appears as white areas. When this noise occurs along the path, it affects the computation of the average gradient, ultimately resulting in a less smooth path. The path generated by NIACO exhibits obvious instability, likely due to its emphasis on real-time performance, with an insufficient number of iterations causing it to remain in a non-converged state. Similar phenomena can be observed in the other sub-figures of Figure 3: the 3D RRT-Connect algorithm fails to find the optimal path; the 3D A* path continues to exhibit insufficient smoothness; and the NIACO path shows significant jitter, with overall poor path quality.

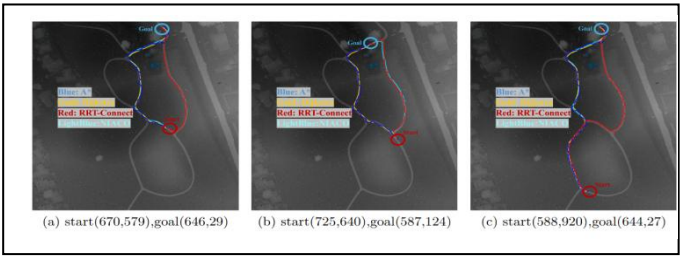

Figure 3. Comparison of 3D Path Planning Algorithms

TABLE II. PERFORMANCE COMPARISON OF 3D PATH PLANNING ALGORITHMS

| Metric | Algorithm | Path 1 | Path 2 | Path 3 |
|---|---|---|---|---|
| Path Cost | A* | 976.8 | 7185.8 | 2161.2 |
|  | Dijkstra | 805.6 | 6185.8 | 1704.2 |
|  | RRT-Connect | 42509.8 | 44331.4 | 58756.0 |
|  | NIACO | 2027.0 | 106594.6 | 15586.2 |
| Computation Time (s) | A* | 153.51 | 191.79 | 188.01 |
|  | Dijkstra | 7.96 | 6.69 | 8.78 |
|  | RRT-Connect | 0.80 | 1.72 | 4.42 |
|  | NIACO | 139.36 | 87.12 | 260.53 |
| Memory Usage (MB) | A* | 35.28 | 35.30 | 35.33 |
|  | Dijkstra | 14.94 | 14.44 | 15.81 |
|  | RRT-Connect | 3.15 | 3.35 | 3.26 |
|  | NIACO | 13.17 | 12.73 | 16.58 |

From the comparison results in Figure 3 and Table 2, it is clear that the 3D A* and 3D Dijkstra algorithms are more advantageous in terms of the resulting path quality. However, the 3D A* algorithm recalculates the average gradient of the surrounding neighborhood every time a pixel is visited, greatly increasing the computational load and resulting in significant time consumption. Although the 3D RRT-Connect algorithm cannot find the optimal path, it benefits from its fast sampling characteristics, which result in very low memory usage. When the step size is appropriately set (3 pixels in this study), the algorithm can find a feasible path in a relatively short amount of time. Due to insufficient iterations, NIACO fails to converge, leading to a high path cost and serious time consumption during iterations. In contrast, the 3D Dijkstra algorithm demonstrates the most stable and excellent performance: it almost always finds the optimal path, requires less computation time, and maintains memory consumption within an acceptable range.

Among the tested methods, 3D Dijkstra demonstrates the most consistent and reliable performance. It successfully finds the optimal path, maintains short computation time, and uses memory resources efficiently. This makes it the most stable and effective algorithm in the current 3D planning context.

## V. CONCLUSION

In summary, this work conducts a comparative evaluation of four path planning algorithms on high-resolution 2D and 3D geospatial road-maps constructed from satellite imagery and airborne LiDAR data, addressing key challenges in autonomous navigation within unstructured environments. In our experimental setup, Dijkstra generally demonstrated stable and balanced performance in terms of path cost, computation time, and memory usage, particularly in static, terrain-aware scenarios. However, it is important to note that performance may vary under different environmental conditions or data configurations. These findings offer insights into the behavior of various planning algorithms when applied to dense geospatial data and support further exploration into adaptive, real-time navigation in more dynamic and unpredictable settings.

ACKNOWLEDGMENT

This work was supported by the National Laboratory of Cooperative Technologies, funded by the Ministry of Culture and Innovation through the National Research, Development and Innovation Fund, under the 2022-2.1.1-NL Establishment and Complex Development of National Laboratories call. Additional support was provided by the Hungarian National Science Foundation (NKFIH OTKA), Grant No. K139485. The source code and sample datasets used in this study can be accessed at: https://github.com/kuma990122/Comparison-of-Path-Planning-Algorithm-for-Autonomous-Vehicle-Using-Satellite-and-Airborne-LiDAR-Data